\documentclass[twoside]{article}
\usepackage[accepted]{arxiv}     
\usepackage[round]{natbib}

\usepackage{times}         
\usepackage{algorithm}         
\usepackage{algorithmic}       
\usepackage{amsfonts}          
\usepackage{amsmath}           
\usepackage{amssymb}
\usepackage{amsopn}            
\usepackage{color}             

\usepackage{graphicx}          

\usepackage{subfig}
\DeclareCaptionType{copyrightbox}

\DeclareMathOperator*{\var}{Var}
\newcommand{\expect}{\mathbb{E}}

\DeclareMathOperator*{\cov}{Cov}

\newcommand{\pmf}{{p}}
\newcommand{\pmfProposal}{{q}}
\newcommand{\pmfOne}{{p_{\rm ini}}}
\newcommand{\pmfTwo}{{p_{\rm tgt}}}
\newcommand{\pmfUnnorm}{{f}}
\newcommand{\pmfUnnormOne}{{f_{\rm ini}}}
\newcommand{\pmfUnnormTwo}{{f_{\rm tgt}}}
\newcommand{\Z}{\mathcal{Z}}
\newcommand{\pfn}{{\Z}}
\newcommand{\pfnOne}{{\Z_{\rm ini}}}
\newcommand{\pfnTwo}{{\Z_{\rm tgt}}}
\newcommand{\pfnTwoEstimate}{{\hat{\Z}_{\rm tgt}}}

\newcommand{\state}{{\bf x}}
\newcommand{\stateSpace}{\mathcal{X}}
\newcommand{\aisWeight}{{w}}
\newcommand{\sampleIdx}{{i}}
\newcommand{\distIdx}{{k}}
\newcommand{\trans}{{T}}
\newcommand{\transReverse}{{\tilde{T}}}

\newcommand{\pmfDirected}{p}

\newcommand{\nsamp}{{M}}
\newcommand{\ndist}{{K}}

\newcommand{\given}{{\,|\,}}
\newcommand{\prob}{{\rm Pr}}

\newcommand{\pathParam}{{\beta}}

\newcommand{\visUnit}{v}
\newcommand{\hidUnit}{h}
\newcommand{\visUnits}{{\bf v}}
\newcommand{\hidUnits}{{\bf h}}
\newcommand{\testVis}{\visUnits_{\rm test}}
\newcommand{\visBiases}{{\bf a}}
\newcommand{\hidBiases}{{\bf b}}
\newcommand{\weights}{{\bf W}}
\newcommand{\nvis}{N_v}
\newcommand{\nhid}{N_h}

\newcommand{\hidUnitsOne}{\hidUnits_1}
\newcommand{\hidUnitsTwo}{\hidUnits_2}
\newcommand{\hidBiasesOne}{\hidBiases_1}
\newcommand{\hidBiasesTwo}{\hidBiases_2}
\newcommand{\weightsOne}{\weights_1}
\newcommand{\weightsTwo}{\weights_2}

\newcommand{\visTrainI}[1]{\visUnits_{\rm tr}^{(#1)}}
\newcommand{\visTestI}[1]{\visUnits_{\rm test}^{(#1)}}
\newcommand{\ntrain}{M_{\rm tr}}
\newcommand{\ntest}{M_{\rm test}}

\newcommand{\pmfForward}{\pmf_{\rm fwd}}
\newcommand{\pmfAnn}{\pmf_{\rm ann}}
\newcommand{\pmfAnnEstimate}{\hat{\pmf}_{\rm ann}}
\newcommand{\pmfProposalForward}{\pmfProposal_{\rm fwd}}
\newcommand{\pmfReverseUnnorm}{\pmfUnnorm_{\rm rev}}
\newcommand{\pmfProposalReverse}{\pmfProposal_{\rm rev}}
\newcommand{\raiseWeight}{w}

\newcommand{\transFrozen}[1]{\trans^{(#1)}}
\newcommand{\transReverseFrozen}[1]{\transReverse^{(#1)}}

\newcommand{\variance}[1]{\var\left[#1\right]}
\newcommand{\covariance}[2]{\cov\left[#1,#2\right]}
\newcommand{\expectation}[1]{\expect\left[#1\right]}

\begin{document}

%

%


\twocolumn[

\runningtitle{Accurate and Conservative Estimates of MRF Log-likelihood using Reverse Annealing}

\aistatstitle{Accurate and Conservative Estimates of MRF Log-likelihood\\
  using Reverse Annealing}

\aistatsauthor{ Yuri Burda\footnotemark[1] \And Roger B. Grosse\footnotemark[1] \And Ruslan Salakhutdinov }

\aistatsaddress{ Fields Institute \\ University of Toronto \And Department of Computer Science \\ University of Toronto \And Department of Computer Science \\ University of Toronto }
 ]

\begin{abstract}
Markov random fields (MRFs) are difficult to evaluate as generative models because computing 
the test log-probabilities requires the intractable partition function.
Annealed importance sampling (AIS) is widely used to estimate MRF partition functions, and often yields quite accurate results. 
However, AIS is prone to overestimate the log-likelihood with little indication that anything is wrong. 
We present the Reverse AIS Estimator (RAISE), a stochastic lower bound on the log-likelihood of an approximation to the original MRF model.
RAISE requires only the same MCMC transition operators as standard AIS.
Experimental results indicate that RAISE agrees closely with AIS log-probability estimates for RBMs, DBMs, and DBNs, but typically errs on the side of underestimating, rather than overestimating, the log-likelihood.
\end{abstract}

\section{Introduction}
In recent years, there has been a resurgence of interest in learning deep representations due to the impressive performance of deep neural networks across a range of tasks. Generative modeling is an appealing method of learning representations, partly because one can directly evaluate a model by measuring the probability it assigns to held-out test data. Restricted Boltzmann machines \citep[RBMs;][]{rbm} and deep Boltzmann machines \citep[DBMs;][]{dbm} are highly effective at modeling various complex visual datasets \citep[\emph{e.g.}][]{ais-rbm,dbm}. Unfortunately, measuring their likelihood exactly is intractable because it requires computing the partition function of a Markov random field (MRF).

Annealed importance sampling \citep[AIS;][]{ais} has emerged as the state-of-the-art algorithm for estimating MRF partition functions, and is widely used to evaluate MRFs as generative models \citep{ais-rbm,deep-not-enough}. AIS is a consistent estimator of the partition function \citep{ais}, and often performs very well in practice. However, it has a property which makes it unreliable: it tends to underestimate the partition function, which leads to overly optimistic measures of the model likelihood. In some cases, it can overestimate the log-likelihood by tens of 
nats \citep[\emph{e.g.}][]{moment-averaging}, and one cannot 
be sure whether impressive test log-probabilities result from a good model or a bad partition function estimator. 
The difficulty of evaluating likelihoods has led researchers to propose alternative generative models for which the log-likelihood can be computed exactly \citep{nade,sum-product} or lower bounded \citep{darn,nvil}, but RBMs and DBMs remain the state-of-the-art for modeling complex data distributions.

\footnotetext[1]{Authors contributed equally}
\setcounter{footnote}{1}

\citet{bounding-test-loglik} highlighted the problem of optimistic RBM log-likelihood estimates and proposed a pessimistic estimator based on nonparametric density estimation. Unfortunately, they reported that their method tends to underestimate log-likelihoods by tens of nats on standard benchmarks, which is insufficient accuracy since the difference between competing models is often on the order of one nat. 

We introduce the Reverse AIS Estimator (RAISE), an algorithm which computes conservative estimates of MRF log-likelihoods, but which achieves similar accuracy to AIS in practice. In particular, consider an approximate generative model defined as the distribution of approximate samples computed by AIS. Using importance sampling with a carefully chosen proposal distribution, RAISE computes a stochastic lower bound on the log-likelihood of the approximate model.  RAISE is simple to implement, as it requires only the same MCMC transition operators as standard AIS.

We evaluated RAISE by using it to estimate test log-probabilities of several RBMs, DBMs, and 
Deep Belief Networks (DBNs). 
The RAISE estimates agree closely with the true log-probabilities on small RBMs where the partition function can be computed exactly. Furthermore, they agree closely with the standard AIS estimates for full-size RBMs, DBMs, and DBNs. Since one estimate is optimistic and one is pessimistic, this agreement is an encouraging sign that both estimates are close to the correct value. Our results suggest that AIS and RAISE, used in conjunction, can provide a practical way of estimating MRF test log-probabilities.

\section{Background}
\vspace{-0.05in}

\subsection{Restricted Boltzmann Machines}
\label{sec:background-rbm}
\vspace{-0.05in}

While our proposed method applies to general MRFs, we use as our running example a particular type of MRF called the restricted Boltzmann machine \citep[RBM;][]{rbm}. An RBM is an MRF with a bipartite structure over a set of visible units $\visUnits = (\visUnit_1, \ldots, \visUnit_{\nvis})$ and hidden units $\hidUnits = (\hidUnit_1, \ldots, \hidUnit_{\nhid})$. In this paper, for purposes of exposition, we assume that all of the variables are binary valued. In this case, the distribution over the joint state $\{\visUnits, \hidUnits\}$ can be written as $\pmfUnnorm(\visUnits, \hidUnits) / \pfn$, where
\begin{equation}
\pmfUnnorm(\visUnits, \hidUnits) = \exp \left( \visBiases^\top \visUnits + \hidBiases^\top \hidUnits + \visUnits^\top \weights \hidUnits \right),
\end{equation}
and $\visBiases$, $\hidBiases$, and $\weights$ denote the visible biases, hidden biases, and weights, respectively. The weights and biases are the RBM's trainable parameters.

To train the RBM's weights and biases, one can
maximize the log-probability of a set of training examples $\visTrainI{1}, \ldots, \visTrainI{\ntrain}$. Since the log-likelihood gradient is intractable to compute exactly, it is typically approximated using contrastive divergence \citep{contrastive-divergence} or persistent contrastive divergence \citep{pcd}. The performance of the RBM is then measured 
in terms of the average log-probability of a set of test examples $\visTestI{1}, \ldots, \visTestI{\ntest}$. 

It remains challenging to evaluate the probability $\pmf(\visUnits) = \pmfUnnorm(\visUnits) / \pfn$ of an example. The unnormalized probability $\pmfUnnorm(\visUnits) = \sum_\hidUnits \pmfUnnorm(\visUnits, \hidUnits)$ can be computed exactly since the conditional distribution factorizes over the $\hidUnit_j$.
However, $\pfn$ is intractable to compute exactly, and must be approximated.

RBMs can also be extended to deep Boltzmann machines \citep{dbm} by adding one or more additional hidden layers. For instance, the joint distribution of a DBM with two hidden layers $\hidUnitsOne$ and $\hidUnitsTwo$ 
can be written as $\pmfUnnorm(\visUnits, \hidUnitsOne, \hidUnitsTwo)/\pfn$, where
\begin{align}
\pmfUnnorm(\visUnits, \hidUnitsOne, \hidUnitsTwo) &= \exp \left( \visBiases^\top \visUnits + \hidBiasesOne^\top \hidUnitsOne + \hidBiasesTwo^\top \hidUnitsTwo + \right. \nonumber \\
&\phantom{=} \left. + \visUnits^\top \weightsOne \hidUnitsOne + \hidUnitsOne^\top \weightsTwo \hidUnitsTwo \right).
\end{align}
DBMs can be evaluated similarly to RBMs. The main difference is that the 
unnormalized probability $\pmfUnnorm(\visUnits) = \sum_{\hidUnitsOne, \hidUnitsTwo} \pmfUnnorm(\visUnits, \hidUnitsOne, \hidUnitsTwo)$ is intractable to compute exactly. However, \citet{dbm} showed that, in practice, the mean-field approximation yields an accurate lower bound. Therefore, similarly to RBMs, the main difficulty in evaluating DBMs is estimating the partition function.

RBMs are also used as building blocks for training Deep Belief Networks \citep[DBNs;][]{dbn}.
For example, a DBN with two hidden layers $\hidUnitsOne$ and $\hidUnitsTwo$ is defined as the probability distribution
\begin{equation}
\pmf(\visUnits, \hidUnitsOne, \hidUnitsTwo) = \pmf_2(\hidUnitsOne,\hidUnitsTwo) \pmf_1(\visUnits \given \hidUnitsOne),
\end{equation}
where $\pmf_2(\hidUnitsOne, \hidUnitsTwo)$ is the probability distribution of an RBM, and $\pmf_1(\visUnits \given \hidUnitsOne)$ is a product of independent logistic units. The unnormalized probability $\pmfUnnorm(\visUnits) = \sum_{\hidUnitsOne,\hidUnitsTwo}{\pmf_1(\visUnits \given \hidUnitsOne)\pmfUnnorm_2(\hidUnitsOne,\hidUnitsTwo)}$ cannot be computed analytically, but can be approximated using importance sampling or a variational lower bound that utilizes a recognition distribution $q(\hidUnitsOne \given \visUnits)$ approximating the posterior $\pmf(\hidUnitsOne \given \visUnits)$ \citep{dbn}.

\subsection{Partition Function Estimation}
\label{sec:background-pfn}
\vspace{-0.05in}

Often we have a probability distribution $\pmfTwo(\state) = \pmfUnnormTwo(\state) / \pfnTwo$ (which we call the \emph{target distribution}) defined on a space $\stateSpace$, where $\pmfUnnormTwo(\state)$ can be computed efficiently for a given $\state \in \stateSpace$, and $\pfnTwo$ is an intractable normalizing constant. There are two particular cases which concern us here. First, $\pmfTwo$ may correspond to a Markov random field (MRF), such as an RBM, where $\pmfUnnormTwo(\state)$ denotes the product of all potentials, and $\pfnTwo = \sum_\state \pmfUnnormTwo(\state)$ is the partition function of the graphical model. 

The second case is where one has a directed graphical model with latent variables $\hidUnits$ and observed variables $\visUnits$. Here, the joint distribution $\pmfDirected(\hidUnits, \visUnits) = \pmfDirected(\hidUnits) \pmfDirected(\visUnits \given \hidUnits)$ can be tractably computed for any particular pair $(\hidUnits, \visUnits)$. However, one often wants to compute the likelihood of a test example $\pmfDirected(\testVis) = \sum_\hidUnits \pmfDirected(\hidUnits, \testVis)$. 
This can be placed in the above 
framework with 
\begin{equation}
\label{eq:marginal}
\pmfUnnormTwo(\hidUnits) = \pmfDirected(\hidUnits) \pmfDirected(\testVis \given \hidUnits)
\hspace{0.1in} \textrm{and} \hspace{0.1in} \pfnTwo = \pmfDirected(\testVis). 
\end{equation}

Mathematically, the two partition function estimation problems outlined above are closely related, and the same classes of algorithms are applicable to each. However, they differ in terms of the behavior of approximate inference algorithms in the context of model selection. 
In particular, many algorithms, such as annealed importance sampling \citep{ais} and sequential Monte Carlo \citep{smc}, yield unbiased estimates $\pfnTwoEstimate$ of the partition function, \emph{i.e.}~$\expect[\pfnTwoEstimate] = \pfnTwo$. Jensen's Inequality shows that such an estimator tends to underestimate the log partition function on average:
\begin{equation}
\expect[\log \pfnTwoEstimate] \leq \log \expect[\pfnTwoEstimate] = \log \pfnTwo.
\end{equation}
In addition, Markov's inequality shows that it is  unlikely to substantially overestimate $\log \pfnTwo$:
\begin{equation}
\prob(\log \pfnTwoEstimate > \log \pfnTwo + b) < e^{-b}. \label{eqn:tail-bound}
\end{equation}
For these reasons, we will refer to the estimator as a \emph{stochastic lower bound} on $\log \pfnTwo$.

In the MRF situation, $\pfnTwo$ appears in the denominator, so underestimates of the log partition function translate into overestimates of the log-likelihood. This is problematic, since inaccurate partition function estimates can lead one to dramatically overestimate the performance of one's model. This problem has led researchers to consider alternative generative models where the likelihood can be tractably computed. By contrast, in the directed case, the partition function is the test log-probability (\ref{eq:marginal}), so underestimates correspond to overly conservative measures of performance. For example, the fact that sigmoid belief networks \citep{deep-sigmoid} have tractable lower (rather than upper) bounds is commonly cited as a reason to prefer them over RBMs and DBMs \citep[\emph{e.g.}][]{nvil}.

We note that it is possible to achieve stronger tail bounds than (\ref{eqn:tail-bound}) by combining multiple unbiased estimates in clever ways \citep{lower-bounding-evidence}.

\subsection{Annealed Importance Sampling}
\label{sec:background-ais}
\vspace{-0.05in}

Annealed importance sampling (AIS) is an algorithm which estimates $\pfnTwo$ by gradually changing, 
or ``annealing,'' a distribution. 
In particular, one must specify a sequence of $\ndist + 1$ intermediate distributions 
$\pmf_\distIdx(\state) = \pmfUnnorm_\distIdx(\state)/\pfn_\distIdx$ for $\distIdx = 0, \ldots \ndist$, 
where $\pmfOne(\state) = \pmf_0(\state)$ is a tractable initial distribution,
and $\pmfTwo(\state) = \pmf_\ndist(\state)$ is the intractable target distribution. 
For simplicity, assume all distributions are strictly positive on $\stateSpace$. 
For each $\pmf_\distIdx$, one must also specify an MCMC transition operator $\trans_\distIdx$ (e.g.~Gibbs sampling) 
which leaves $\pmf_\distIdx$ invariant. AIS alternates between MCMC transitions and importance sampling updates, as shown in Algorithm~\ref{alg:ais}.

\begin{figure}[t]
\vspace{-0.2in}
\begin{minipage}[t]{0.5\textwidth}
\begin{algorithm}[H]
\caption{Annealed Importance Sampling}
\label{alg:ais}
\begin{small}
\begin{algorithmic}
	\FOR{$\sampleIdx = 1 \textrm{ to } \nsamp$} 
		\STATE $\state_0 \gets$ sample from $\pmf_0(\state) = \pmfUnnorm_{\textrm{ini}}(\state)/\pfnOne$
		\STATE $\aisWeight^{(\sampleIdx)} \gets \pfnOne$
		\FOR{$\distIdx = 1 \textrm{ to } \ndist$} 
			\STATE $\aisWeight^{(\sampleIdx)} \gets \aisWeight^{(\sampleIdx)} \frac{\pmfUnnorm_{\distIdx}(\state_{\distIdx-1})}{\pmfUnnorm_{\distIdx - 1}(\state_{\distIdx-1})}$
			\STATE $\state_\distIdx \gets$ sample from $\trans_\distIdx\left(\cdot \given \state_{\distIdx-1}\right)$
		\ENDFOR
	\ENDFOR
	\RETURN $\pfnTwoEstimate = \sum_{\sampleIdx=1}^\nsamp \aisWeight^{(\sampleIdx)}/\nsamp$
\end{algorithmic}
\end{small}
\end{algorithm}
\end{minipage}
\vspace{-0.1in}
\end{figure}

The output of AIS is an unbiased estimate \smash{$\pfnTwoEstimate$} of $\pfnTwo$. Importantly, unbiasedness is not an asymptotic property, but holds for any $\ndist$ \citep{ais, jarzynski97b}. \citet{ais} demonstrated this by viewing AIS as an importance sampling estimator over an extended state space. In particular, define the distributions 
\begin{align}
\pmfProposalForward(\state_{0:\ndist-1}) &= \pmf_0(\state_0) \prod_{\distIdx=1}^{\ndist-1} \trans_\distIdx(\state_{\distIdx} \given \state_{\distIdx - 1})  \\
\pmfReverseUnnorm(\state_{0:\ndist-1}) &= \pmfUnnormTwo(\state_{\ndist-1}) \prod_{\distIdx=1}^{\ndist-1} \transReverse_\distIdx(\state_{\distIdx-1} \given \state_\distIdx),
\end{align}
where $\transReverse_\distIdx(\state^\prime \given \state) = \trans_\distIdx(\state \given \state^\prime) \pmf_\distIdx(\state^\prime) / \pmf_\distIdx(\state)$ is the reverse transition operator for $\trans_\distIdx$.
Here, $\pmfProposalForward$ represents the sequence of states generated by AIS, and $\pmfReverseUnnorm$ is a fictitious (unnormalized) reverse chain which begins with an exact sample from $\pmfTwo$ and applies the transitions in reverse order. 
\citet{ais} showed that the AIS weights correspond to the importance weights for $\pmfReverseUnnorm$ with $\pmfProposalForward$ as the proposal distribution.

The mathematical formulation of AIS leaves much flexibility for choosing intermediate distributions. The choice of distributions can have a large effect on the performance of AIS \citep{moment-averaging}, but the most common choice is to take geometric averages of the initial and target distributions:
\begin{equation}
   \hspace{-0.04in} p_{\beta}(\state) = \pmfUnnorm_\pathParam(\state)/\Z(\beta) = \pmfUnnormOne(\state)^{1 - \pathParam} \pmfUnnormTwo(\state)^\pathParam / \Z(\beta), \hspace{-0.02in} \label{eqn:geometric-averages}
\end{equation}
where $0 = \beta_0 < \beta_1 < ... < \beta_K = 1$ defines the annealing schedule.  
Commonly, $\pmfUnnormOne$ is the uniform distribution, and~(\ref{eqn:geometric-averages}) reduces to 
$p_{\beta}(\state) = \pmfUnnormTwo(\state)^\pathParam/\Z(\beta)$. This motivates the term ``annealing'', and $\pathParam$ resembles an inverse temperature parameter. As in simulated annealing, the ``hotter'' distributions often allow faster mixing between modes which are isolated in $\pmfTwo$. Geometric averages are widely used because they often have a simple form; for instance, the geometric average of two RBMs is obtained by linearly averaging the weights and biases. The values of $\beta$ can be spaced evenly between 0 and 1, although other schedules have been explored \citep{neal96,behrens12,calderhead09}. 

\section{Reverse AIS Estimator}
\label{sec:raise}
\vspace{-0.1in}
A significant difficulty in evaluating MRFs is that it is intractable to compute the partition function. Furthermore, the commonly used algorithms, such as AIS, tend to \emph{overestimate} the log-likelihood. If we cannot hope to obtain provably accurate partition function estimates, it would be far preferable for algorithms to \emph{underestimate}, rather than \emph{overestimate}, the log-likelihoods. This would save us from the embarrassment of reporting 
unrealistically high test log-probability scores for a given dataset. In this section, we define an approximate generative model which becomes equivalent to the MRF in the limit of infinite computation. We then present a procedure for obtaining unbiased 
estimates of the probability of a test example
(and therefore a stochastic lower bound on the test log-probability) under the approximate model.

\subsection{Case of Tractable Posterior}
\label{sec:raise-tractable}

\vspace{-0.05in}
In this section, we denote the model state as $\state = (\visUnits, \hidUnits)$, with $\visUnits$ observed and $\hidUnits$ unobserved. 
Let us first assume the conditional distribution $\pmfTwo(\hidUnits \given \visUnits)$ is tractable, as is the case for RBMs. 
Define the following generative process, which corresponds to the sequence of transitions in AIS:
\begin{equation}
\pmfForward(\state_{0:\ndist}) = \pmf_0(\state_0) \prod_{\distIdx=1}^{\ndist} \trans_\distIdx(\state_\distIdx \given \state_{\distIdx-1}).
\end{equation}
By taking the final visible states of this process, we obtain a generative model (which we term the \emph{annealing model}) which approximates $\pmfTwo(\visUnits)$:
\begin{equation}
\pmfAnn(\visUnits_\ndist) = \sum_{\state_{0:\ndist-1}, \hidUnits_\ndist} \pmfForward(\state_{0:\ndist-1}, \hidUnits_\ndist, \visUnits_\ndist). \label{eq:raise-model}
\end{equation}

\begin{figure}[t]
\vspace{-0.2in}
\begin{minipage}[t]{0.5\textwidth}
\begin{algorithm}[H]
\caption{Reverse AIS Estimator (RAISE)}
\label{alg:raise}
\begin{small}
\begin{algorithmic}
        \FOR{$\sampleIdx = 1 \textrm{ to } \nsamp$}
                \STATE $\hidUnits_\ndist \gets \textrm{sample from } \pmfTwo(\hidUnits \given \testVis)$
                \STATE $\aisWeight^{(\sampleIdx)} \gets \pmfUnnormTwo(\testVis) / \pfn_0$
                \FOR{$\distIdx = \ndist - 1 \textrm{ to } 0$}
                        \STATE $\state_\distIdx \gets$ sample from $\transReverse_\distIdx \left(\cdot \given \state_{\distIdx+1}\right)$
                        \STATE $\aisWeight^{(\sampleIdx)} \gets \aisWeight^{(\sampleIdx)} \frac{\pmfUnnorm_{\distIdx}(\state_\distIdx)}{\pmfUnnorm_{\distIdx + 1}(\state_\distIdx)}$
                \ENDFOR
        \ENDFOR
        \RETURN $\pmfAnnEstimate(\testVis) = \sum_{\sampleIdx=1}^\nsamp \raiseWeight^{(\sampleIdx)}/\nsamp$
\end{algorithmic}
\end{small}
\end{algorithm}
\end{minipage}
\vspace{-0.1in}
\end{figure}

Suppose we are interested in estimating the probability of a test example $\testVis$. We use as a proposal distribution a reverse chain starting from $\testVis$. In the annealing metaphor, this corresponds to gradually ``melting'' the distribution:
\begin{equation}
\hspace{-0.1in}
\pmfProposalReverse(\state_{0:\ndist-1}, \hidUnits_\ndist \given \testVis) = \pmfTwo(\hidUnits_\ndist \given \testVis) \prod_{\distIdx=1}^{\ndist} \transReverse_\distIdx(\state_{\distIdx-1} \given \state_\distIdx), \nonumber 
\end{equation}
where we identify $\visUnits_\distIdx=\testVis$, and $\transReverse_\distIdx(\state^\prime \given \state) = \trans_\distIdx(\state \given \state^\prime) \pmf_\distIdx(\state^\prime) / \pmf_\distIdx(\state)$ is the reverse transition operator for $\trans_\distIdx$. We then obtain the following identity:
\begin{small}
\begin{align}
\pmfAnn(\testVis) &= \expect_{\pmfProposalReverse} \left[ \frac{\pmfForward(\state_{0:\ndist-1}, \hidUnits_\ndist, \testVis)}{\pmfProposalReverse(\state_{0:\ndist-1}, \hidUnits_\ndist \given \testVis)} \right] \nonumber \\
&= \expect_{\pmfProposalReverse} \left[  \frac{\pmf_0(\state_0)}{\pmfTwo(\hidUnits_\ndist \given \testVis)} \prod_{\distIdx=1}^\ndist \frac{\trans_\distIdx(\state_\distIdx \given \state_{\distIdx-1})}{\transReverse_\distIdx(\state_{\distIdx-1} \given \state_\distIdx)} \right] \nonumber \\
&= \expect_{\pmfProposalReverse} \left[  \frac{\pmf_0(\state_0)}{\pmfTwo(\hidUnits_\ndist \given \testVis)} \prod_{\distIdx=1}^\ndist \frac{\pmfUnnorm_\distIdx(\state_\distIdx)}{\pmfUnnorm_\distIdx(\state_{\distIdx-1})} \right] \nonumber \\
&= \expect_{\pmfProposalReverse} \left[  \frac{\pmf_0(\state_0)}{\pmfTwo(\hidUnits_\ndist \given \testVis)} \frac{\pmfUnnormTwo(\state_\ndist)}{\pmfUnnorm_0(\state_0)} \prod_{\distIdx=0}^{\ndist-1} \frac{\pmfUnnorm_\distIdx(\state_\distIdx)}{\pmfUnnorm_{\distIdx+1}(\state_\distIdx)} \right] \nonumber \\
&= \expect_{\pmfProposalReverse} \left[ \frac{\pmfUnnorm_\ndist(\testVis)}{\pfn_0} \prod_{\distIdx=0}^{\ndist-1} \frac{\pmfUnnorm_\distIdx(\state_\distIdx)}{\pmfUnnorm_{\distIdx+1}(\state_\distIdx)} \right] \nonumber \\
&\triangleq \expect_{\pmfProposalReverse} \left[ \raiseWeight \right]. \label{eqn:raise-weight}
\end{align}
\end{small}
This yields the following algorithm: generate $\nsamp$ samples from $\pmfProposalReverse$, and average the values $\raiseWeight$ defined in (\ref{eqn:raise-weight}). There is no need to store the full chains, since the weights can be updated online. We refer to this algorithm as the Reverse AIS Estimator, or RAISE. 
The full algorithm is given in Algorithm \ref{alg:raise}. We note that RAISE is straightforward to implement, as it requires only the same MCMC transition operators as standard AIS.

Our derivation (\ref{eqn:raise-weight}) mirrors the derivation of AIS by \citet{ais}. The difference is that in AIS, the reverse chain is merely hypothetical; in RAISE, the reverse chain is simulated, and it is the forward chain which is hypothetical.

By (\ref{eqn:raise-weight}), the weights $\raiseWeight$ are an unbiased estimator of the probability $\pmfAnn(\testVis)$. Therefore, following the discussion of Section~\ref{sec:background-pfn}, $\log \raiseWeight$ is a stochastic lower bound on $\log \pmfAnn(\testVis)$. Furthermore, since $\pmfAnn$ converges to $\pmfTwo$ in probability as $\ndist \rightarrow \infty$ \citep{ais}, we would heuristically expect RAISE to yield a conservative estimate of $\log \pmfTwo(\testVis)$. This is not strictly guaranteed, however; RAISE may overestimate $\log \pmfTwo(\testVis)$ for finite $\ndist$ if $\pmfAnn(\testVis) > \pmfTwo(\testVis)$, which is possible if the AIS approximation somehow attenuates pathologies in the original MRF. (One such example is described in Section~\ref{sec:rbm-experiments}.) However, since RAISE is a stochastic lower bound on the log-probabilities under the annealing model, we can strictly rule out the possibility of RAISE reporting unrealistically high test log-probabilities for a given dataset, a situation frequently observed with AIS.

\begin{figure*}
\vspace{-0.1in}
\begin{center}
\includegraphics[width=\textwidth]{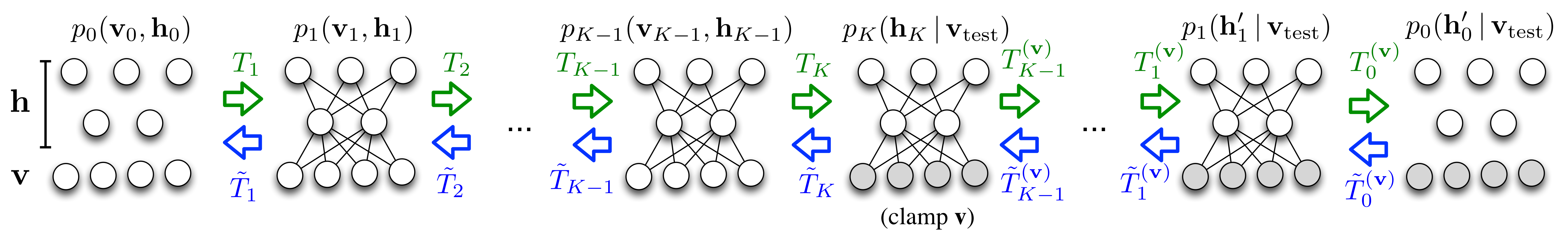}
\end{center}
\vspace{-0.2in}
\caption{\small A schematic of RAISE for intractable distributions, applied to DBMs. Green: generative model. Blue: proposal distribution. At the top is shown which distribution the variables at each step are meant to approximate.}
\label{fig:raise-dbm}
\vspace{-0.1in}
\end{figure*}

\subsection{Extension to Intractable Posterior Distributions}
\label{sec:intractable-posterior}
\vspace{-0.05in}
Because Algorithm \ref{alg:raise} begins with an exact sample from the conditional distribution $\pmfTwo(\hidUnits \given \testVis)$, it requires that this distribution be tractable. However, many models of interest, such as DBMs, have intractable posterior distributions. To deal with this case, we augment the forward chain with an additional heating step, such that the conditional distribution in the final step is tractable, but the distribution over $\visUnits$ agrees with that of $\pmfAnn$ in (\ref{eq:raise-model}). We make the further (weak) assumption that $\pmf_0(\hidUnits \given \visUnits)$ is tractable. Let $\transFrozen{\visUnits}_\distIdx$ denote an MCMC transition operator which preserves $\pmf_\distIdx(\visUnits, \hidUnits)$, but does not change $\visUnits$. 
For example, it may cycle through Gibbs updates 
to all variables except $\visUnits$. 
The forward chain then has the following distribution:
\begin{align}
\pmfForward(\state_{0:\ndist}, \hidUnits^\prime_{0:\ndist-1}) &= \pmf_0(\state_0) \prod_{\distIdx=1}^{\ndist} \trans_\distIdx(\state_\distIdx \given \state_{\distIdx-1}) \nonumber \\
&\phantom{=} \prod_{\distIdx=0}^{\ndist-1} \transFrozen{\visUnits_\ndist}_\distIdx(\hidUnits^\prime_\distIdx \given \hidUnits^\prime_{\distIdx+1}),
\nonumber 
\end{align}
where we identify $\hidUnits^\prime_\ndist = \hidUnits_\ndist$. The reverse distribution is given by:
\begin{align}
&\pmfProposalReverse(\state_{0:\ndist-1}, \hidUnits_\ndist, \hidUnits^\prime_{0:\ndist-1} \given \testVis) = \nonumber \\
& \pmf_0(\hidUnits^\prime_0 \given \testVis) \prod_{\distIdx=0}^{\ndist-1} \transReverseFrozen{\testVis}_{\distIdx}(\hidUnits^\prime_{\distIdx+1} \given \hidUnits^\prime_{\distIdx}) \prod_{\distIdx=1}^{\ndist} \transReverse_\distIdx(\state_{\distIdx-1} \given \state_\distIdx).
\nonumber 
\end{align}
The unbiased estimator is derived similarly to that of Section~\ref{sec:raise}:
\begin{align}
\raiseWeight &\triangleq \frac{\pmfForward(\state_{0:\ndist-1}, \hidUnits_\ndist, \testVis, \hidUnits^\prime_{0:\ndist-1})}{\pmfProposalReverse(\state_{0:\ndist-1}, \hidUnits_\ndist, \hidUnits^\prime_{0:\ndist-1} \given \testVis)} \\
&= \pmf_0(\testVis) \prod_{\distIdx=0}^{\ndist-1} \frac{\pmfUnnorm_\distIdx(\state_\distIdx)}{\pmfUnnorm_{\distIdx+1}(\state_\distIdx)} \prod_{\distIdx=1}^\ndist \frac{\pmfUnnorm_\distIdx(\hidUnits^\prime_\distIdx, \testVis)}{\pmfUnnorm_{\distIdx-1}(\hidUnits^\prime_\distIdx, \testVis)}
\nonumber 
\end{align}
The full algorithm is shown in Algorithm~\ref{alg:raise-intractable-posterior}, and a schematic for the case of DBMs is shown in Figure~\ref{fig:raise-dbm}. 

\begin{figure}
\vspace{-0.1in}
\begin{minipage}[t]{0.5\textwidth}
\begin{algorithm}[H]
\caption{RAISE with intractable posterior}
\label{alg:raise-intractable-posterior}
\begin{small}
\begin{algorithmic}
	\FOR{$\sampleIdx = 1 \textrm{ to } \nsamp$} 
		\STATE $\hidUnits^\prime_0 \gets \textrm{sample from } \pmf_0(\hidUnits \given \testVis)$
		\STATE $\aisWeight^{(\sampleIdx)} \gets \pmf_0(\testVis)$
		\FOR{$\distIdx = 1 \textrm{ to } \ndist$} 
			\STATE $\hidUnits^\prime_\distIdx \gets$ sample from $\transReverseFrozen{\testVis}_\distIdx \left(\cdot \given \hidUnits^\prime_{\distIdx-1}\right)$
			\STATE $\aisWeight^{(\sampleIdx)} \gets \aisWeight^{(\sampleIdx)} \frac{\pmfUnnorm_{\distIdx}(\hidUnits^\prime_\distIdx, \testVis)}{\pmfUnnorm_{\distIdx - 1}(\hidUnits^\prime_\distIdx, \testVis)}$
		\ENDFOR
		\FOR{$\distIdx = \ndist - 1 \textrm{ to } 0$} 
			\STATE $\state_\distIdx \gets$ sample from $\transReverse_\distIdx \left(\cdot \given \state_{\distIdx+1}\right)$
			\STATE $\aisWeight^{(\sampleIdx)} \gets \aisWeight^{(\sampleIdx)} \frac{\pmfUnnorm_{\distIdx}(\state_\distIdx)}{\pmfUnnorm_{\distIdx + 1}(\state_\distIdx)}$
		\ENDFOR
	\ENDFOR
	\RETURN $\pmfAnnEstimate(\testVis) = \sum_{\sampleIdx=1}^\nsamp \raiseWeight^{(\sampleIdx)}/\nsamp$
\end{algorithmic}
\end{small}
\end{algorithm}
\end{minipage}
\vspace{-0.1in}
\end{figure}

\subsection{Interpretation as Unrolling}
\vspace{-0.05in}
\citet{dbn} showed that the Gibbs sampling procedure for a binary RBM could be interpreted as generating from an infinitely deep sigmoid belief net with shared weights. They used this insight to derive a greedy training procedure for Deep Belief Nets (DBNs), where one unties the weights of a single layer at a time. Furthermore, they observed that one could perform approximate inference in the belief net using the transpose of the generative weights to compute a variational approximation.

We note that, for RBMs, RAISE can similarly be viewed as a form of unrolling: the annealed generative model $\pmfAnn$ can be viewed as a belief net with $K+1$ layers. Furthermore, the RAISE proposal distribution can be viewed as using the transpose of the weights to perform approximate inference. (The difference from approximate inference in DBNs is that RAISE samples the units rather than using the mean-field approximation). 

\begin{figure}
\vspace{-0.1in}
\begin{center}
\includegraphics[width=0.2\textwidth]{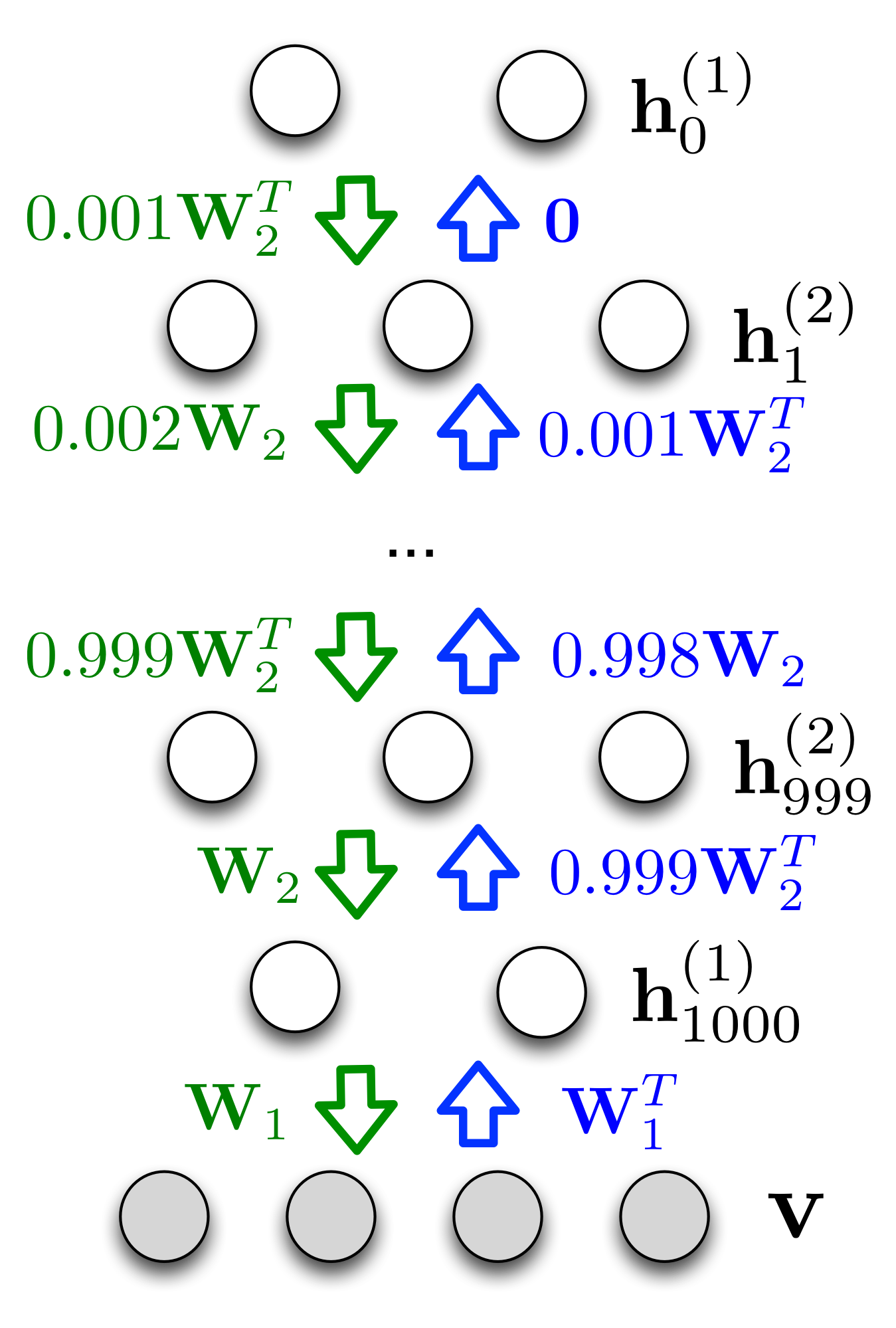}
\end{center}
\vspace{-0.1in}
\caption{\small RAISE applied to a DBN unrolled into a very deep sigmoid belief net, for $\ndist=1000$ intermediate distributions. {\bf Green:} generative model. {\bf Blue:} proposal distribution.}
\label{fig:unroll}
\vspace{-0.1in}
\end{figure}

This interpretation of RAISE suggests a method of applying it to DBNs. The generative model is obtained by unrolling the RBM on top of the directed layers as shown in Figure~\ref{fig:unroll}. 
The proposal distribution uses the transposes of the DBN weights for each of the directed layers.
The rest is the same as the ordinary RAISE for the unrolled part of the model. 

\section{Variance Reduction using Control Variates}
\label{sec:variance-reduction}
\vspace{-0.1in}
One of the virtues of log-likelihood estimation using AIS is its speed: the partition function need only be estimated once. 
RAISE, unfortunately, must be run separately for every test example. 
We would therefore prefer to compute the RAISE estimate for only a small number of test examples. 
Unfortunately, subsampling the test examples introduces a significant source of variability: as different test examples can have wildly different log-likelihoods\footnote{This effect can be counterintuitively large due to different complexities of different categories; \emph{e.g.}, for the mnistCD25-500 RBM, 
the average log-likelihood of handwritten digits ``1'' was 56.6 nats higher than 
the average log-likelihood of digits ``8''.}, the estimate of the average log-likelihood can vary significantly depending which batch of examples is selected. We attenuate this variability using the method of control variates \citep{ross-simulation}, a variance reduction technique which has also been applied to black-box variational inference \citep{black-box-variational}.

If $Y_1, \ldots, Y_n$ are independent samples of a random variable $Y$, then the sample average
$\frac{1}{n}\sum_{i=1}^nY_i$
is an unbiased estimator of $\expectation{Y}$ with variance $\variance{Y}/n$. If $X$ is another random variable (which ideally is both cheap to compute and highly correlated with $Y$), then for any scalar $\alpha$,
\begin{equation}
\frac{1}{n}\sum_{i=1}^n\left(Y_i- \alpha X_i\right)+\frac{\alpha}{N}\sum_{i=1}^NX_i \label{eqn:control-variate}
\end{equation}
is an unbiased estimator of $\expectation{Y}$ with variance $$\frac{\variance{Y- \alpha X}}{n}+ \alpha^2 \frac{\variance{X}}{N}+2\alpha \frac{\covariance{Y-\alpha X}{X}}{n}.$$

In our experiments, $Y$ is the RAISE estimate of the log-probability of a test example, and $X$ is the 
(exact or estimated) log unnormalized probability under the original MRF. Since the unnormalized probability under the MRF is significantly easier to evaluate than the log-probability under the annealing model, we can let $N$ to be much larger than $n$; we set $n=100$ and let $N$ be the total number of test examples. Since the annealing model is an approximation to the MRF, the two models should assign similar log-probabilities, so we set $\alpha = 1$.
Hence we expect the variance of $Y-X$ to be smaller than the variance of $Y$, and thus (\ref{eqn:control-variate}) to have a significantly smaller variance than the sample average. Empirically, we have found that $Y-X$ has significantly smaller variance than $Y$, even when the number of intermediate distributions is relatively small.

\section{Experimental Results}
\vspace{-0.1in}
We have evaluated RAISE on several MRFs to determine if its log-probability estimates are both accurate and conservative.
We compared our estimates against those obtained from standard AIS.
We also compared against the exact log-probabilities of small models for which the partition function can be computed exactly~\citep{ais-rbm}.
AIS is expected to overestimate the true log-probabilities while RAISE is expected 
to underestimate them. Hence, a close agreement between the two estimators would be a strong indication of accurate estimates. 

We considered two datasets: (1) the MNIST handwritten digit dataset~\citep{mnist}, which has long served as a benchmark for both classification and density modeling, and (2) the Omniglot dataset \citep{omniglot}, which contains images of handwritten characters across many world alphabets.\footnote{We used the standard split of MNIST into 60,000 training and 10,000 test examples and a random split of Omniglot into 24,345 training and 8,070 test examples. In both cases, the inputs are $28 \times 28$ binary images.}

Both AIS and RAISE can be used with any sequence of intermediate distributions. For simplicity, in all of our experiments, we used the geometric averages path (\ref{eqn:geometric-averages}) with linear spacing of the parameter $\pathParam$. 
We tested two choices of initial distribution $\pmfOne$: the uniform distribution, and the data base rate (DBR) distribution \citep{ais-rbm}, where all units are independent, all hidden units are uniform, and the visible biases are set to match the average pixel values in the training set. In all cases, our MCMC transition operator was Gibbs sampling.

We estimated the log-probabilities of a random sample of 100 examples from the 
test set using RAISE and used the method of control variates (Sec.~\ref{sec:variance-reduction}) to estimate the average log-probabilities on the full test dataset. 
For RBM experiments, the control variate was the RBM log unnormalized probability, $\log \pmfUnnorm(\visUnits)$, whereas
for DBMs and DBNs, we used an estimate based on simple importance sampling as described below. 
For each of the 100 test examples, RAISE was run with 50 independent chains, while the AIS partition function estimates used 5,000 chains; this closely matched the computation time per intermediate distribution between the two methods. Each method required about 1.5 hours with the largest number of intermediate distributions ($\ndist= \textrm{100,000}$).

\begin{table*}[t]
\vspace{-0.1in}
\begin{center}
\begin{scriptsize}
\begin{tabular}{l|r|r|rrr|rrr}
 \multicolumn{3}{c}{} & \multicolumn{3}{c}{uniform}            & \multicolumn{3}{c}{data base rates} \\
Model         & exact     & CSL      & RAISE    & AIS      & gap    & RAISE    & AIS       & gap        \\ \hline
mnistCD1-20   & -164.50   & -185.74  & -165.33  & -164.51  & 0.82   & -164.11  & -164.50   & -0.39     \\
mnistPCD-20   & -150.11   & -152.13  & -150.58  & -150.04  & 0.54   & -150.17  & -150.10   & 0.07    \\
mnistCD1-500  & ---       & -566.91  & -150.78  & -106.52  & 44.26  & -124.77  & -124.09   & 0.68       \\
mnistPCD-500  & ---       & -138.76  & -101.07  & -99.99   & 1.08   & -101.26  & -101.28   & -0.02    \\
mnistCD25-500 & ---       & -145.26  & -88.51   & -86.42   & 2.09   & -86.39   & -86.35    & 0.04    \\
omniPCD-1000  & ---       & -144.25  & -100.47  & -100.45  & 0.02   & -100.46  & -100.46   & 0.00
\end{tabular}
\end{scriptsize}
\end{center}
\vspace{-0.1in}
\caption{\small RAISE and AIS average test log-probabilities using 100,000 intermediate distributions and both choices of $\pmfOne$. {\bf CSL:} the estimator of \citet{bounding-test-loglik}. {\bf gap:} the difference $\textrm{AIS} - \textrm{RAISE}$}
\label{tbl:raise-vs-ais}
\vspace{-0.1in}
\end{table*}

\subsection{Restricted Boltzmann Machines}
\label{sec:rbm-experiments}
\vspace{-0.05in}
We considered models trained using two algorithms: contrastive divergence \citep[CD;][]{contrastive-divergence} with both 1 and 25 CD steps, and persistent contrastive divergence \citep[PCD;][]{pcd}. We will refer to the RBMs by the dataset, training algorithm, and the number of hidden units.
For example, ``mnistCD25-500'' denotes an RBM with 500 hidden units, trained on MNIST using 25 CD steps. 
The MNIST trained RBMs are the same ones evaluated by \citet{moment-averaging}.
We also provide comparisons to the Conservative Sampling-based Log-likelihood (CSL) estimator of 
\citet{bounding-test-loglik}.\footnote{The number of chains and number of Gibbs steps for CSL were chosen to match the total number of Gibbs steps required by RAISE and AIS for $\ndist= \textrm{100,000}$.}

\begin{figure}[t]
\vspace{-0.1in}
\includegraphics[width=0.4\textwidth]{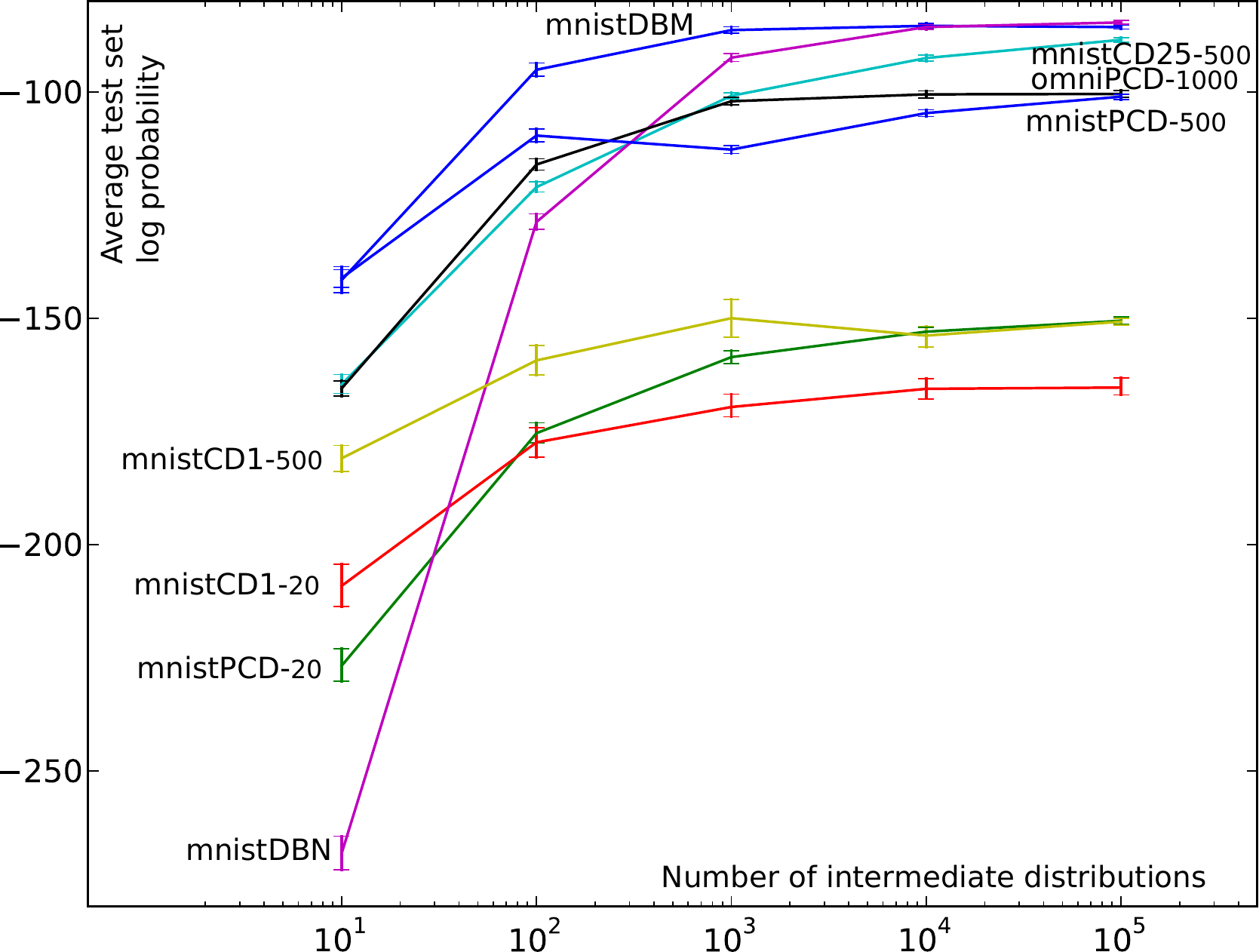}
\caption{\small RAISE estimates of average test log-probabilities using uniform $\pmfOne$. The log-probability estimates tend to increase with the number of intermediate distributions, suggesting that RAISE is a conservative estimator.}
\label{fig:log-likelihood-rbms}
\vspace{-0.2in}
\end{figure}

Figure \ref{fig:log-likelihood-rbms} shows the average RAISE test log-probability estimates for all of the RBMs 
as a function of the number of intermediate distributions. 
In all of these examples, as expected, the estimated log-probabilities tended 
to increase with the number of intermediate distributions, consistent with RAISE being
a conservative log-probability estimator.

Table \ref{tbl:raise-vs-ais} shows the final average test log-probability 
estimates obtained using CSL as well as both RAISE and AIS with 100,000 intermediate distributions. 
In all of the trials using the DBR initial distribution, 
the estimates of AIS and RAISE agreed to within 1 nat, and in many cases, to within 0.1 nats.  
The CSL estimator, on the other hand, underestimated $\log \pmfTwo$ by tens of nats in almost all cases, 
which is insufficient accuracy since well-trained models often differ by only a few nats.

\begin{figure}[t]%
  \subfloat[uniform]{\includegraphics[width=0.48\linewidth]{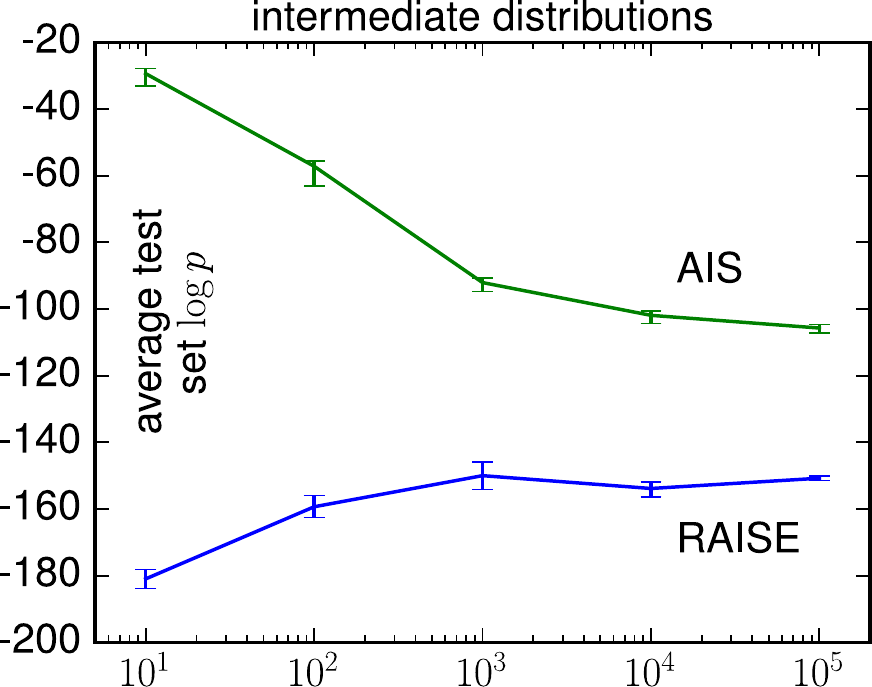}}%
  \hfill
  \subfloat[data base rates]{\includegraphics[width=0.48\linewidth]{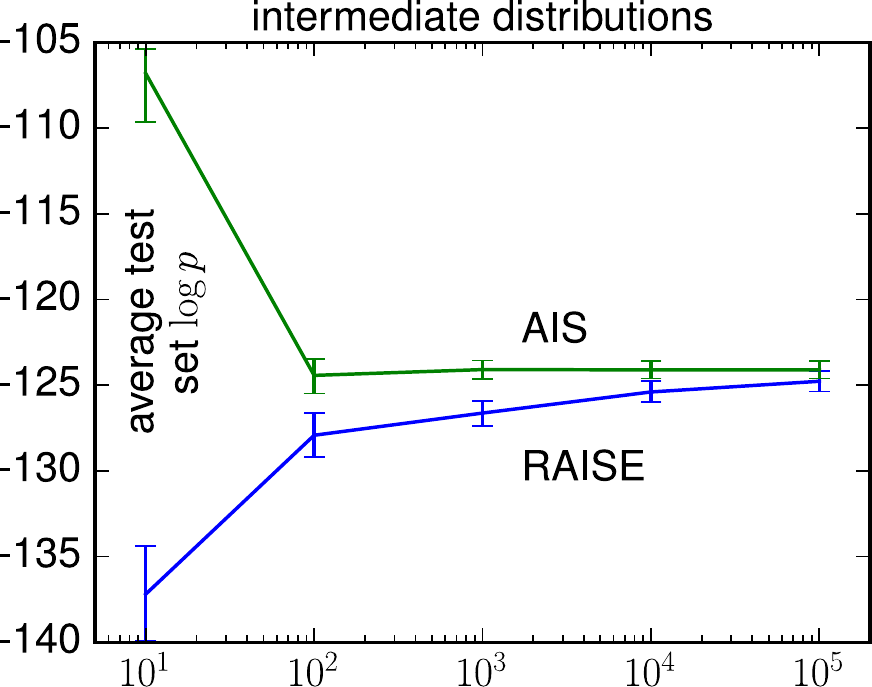}}%
  \caption{\small AIS and RAISE estimates of mnistCD1-500 average test log-probabilities have a significant gap when annealing from a uniform initial distribution. However,
they agree closely when annealing from the data base rates.}
    \label{fig:cd1-500}
\vspace{-0.1in}
\end{figure}

We observed that the DBR initial distribution gave consistently better agreement between the two methods
compared with the uniform distribution, consistent with the results of \citet{ais-rbm}. The largest
discrepancy, 44.26 nats, was for mnistCD1-500 with uniform $\pmfOne$; with DBR, the two methods differed
by only 0.68. Figure~\ref{fig:cd1-500} plots both estimates as a function of the number of initial distributions.
In the uniform case, one might not notice the inaccuracy only by running AIS, as the AIS estimates may
appear to level off. One could be tricked into reporting results that are tens of nats too high!
By contrast, when both methods are run in conjunction, the inaccuracy of at least one
of the methods becomes obvious.

\begin{figure}
\vspace{-0.1in}
\begin{minipage}{0.26 \textwidth}
\includegraphics[width=\textwidth]{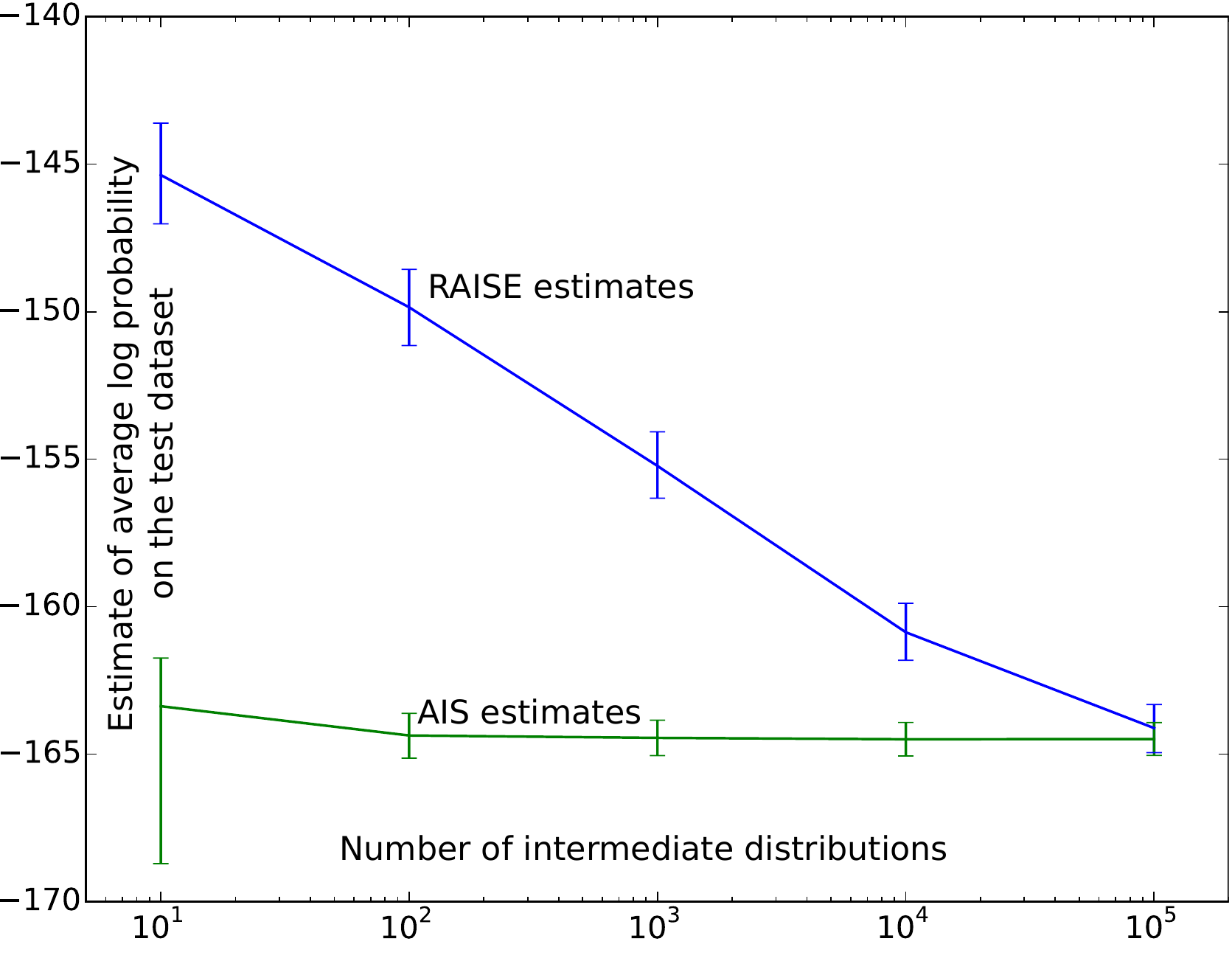}
\end{minipage}
\begin{minipage}{0.2 \textwidth}
\includegraphics[width=\textwidth]{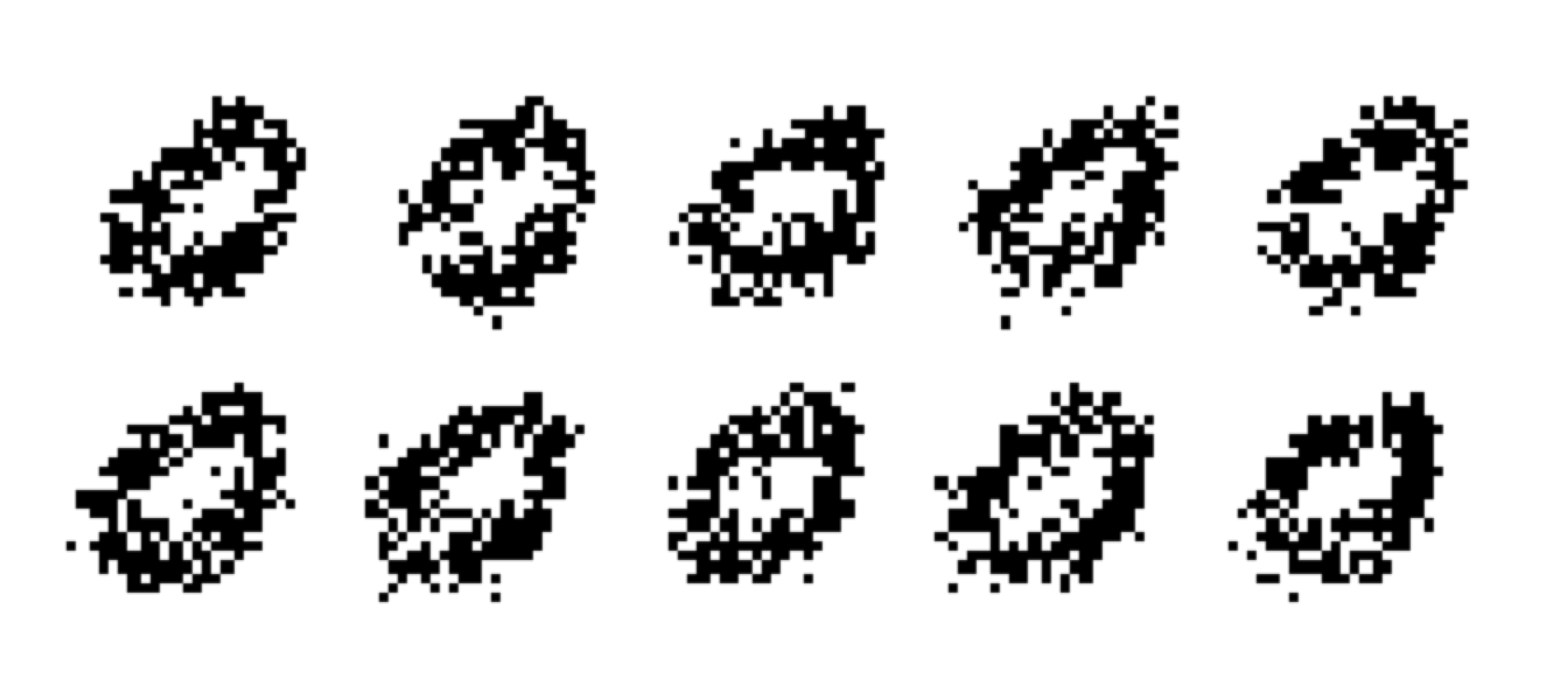} \\
\vspace{1em}
\includegraphics[width=\textwidth]{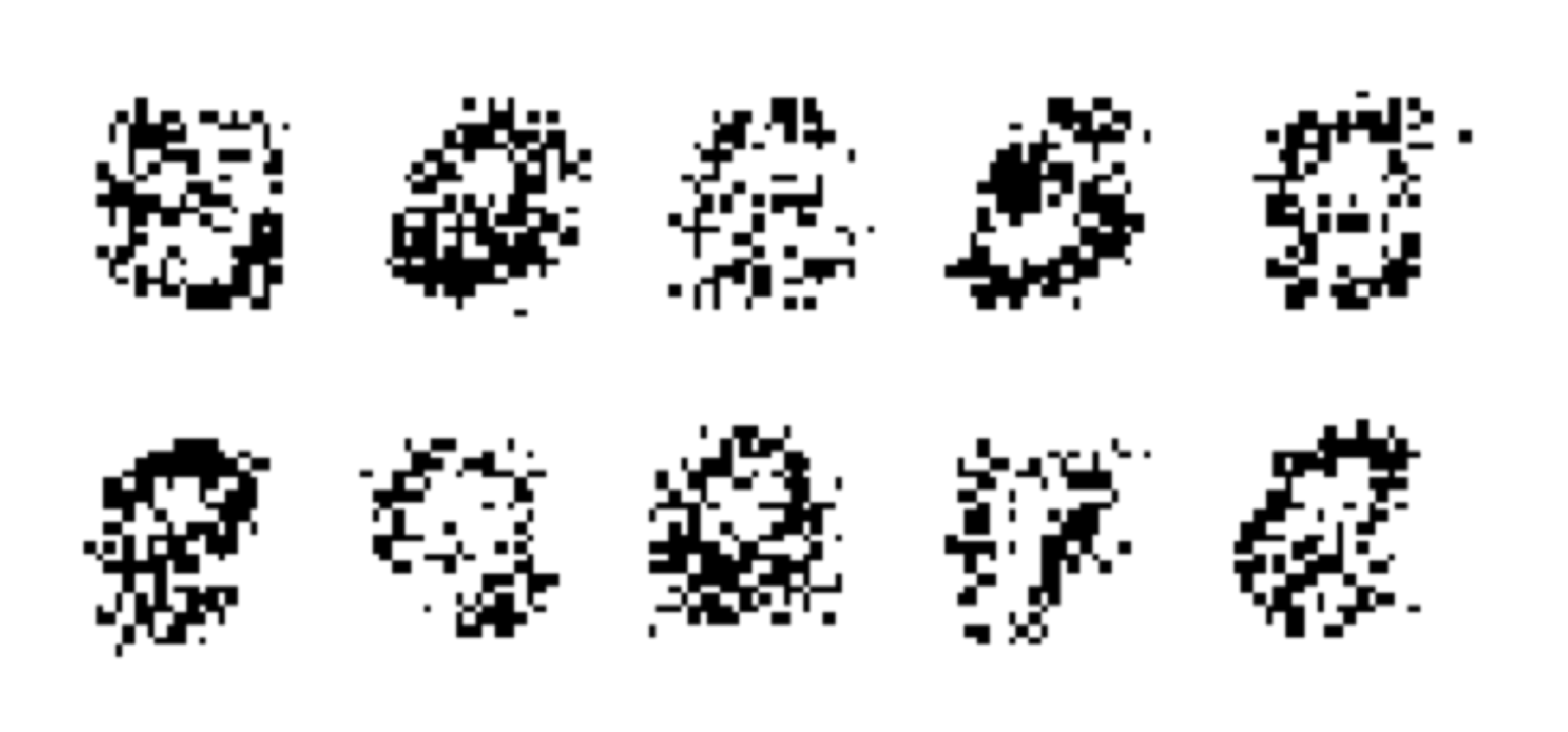}
\end{minipage}
\vspace{-0.1in}
\caption{\small The mnistCD1-20 RBM, where we observed RAISE to overestimate the RBM's
test log-probabilities. {\bf Left:} Average test log-probability estimates as a function of $K$. {\bf Top right:} 10 independent samples from the RBM. {\bf Bottom right:} 10 independent samples from the annealing model $\pmfAnn$ with 10 intermediate distributions. The $\pmfAnn$ samples, while poor, show greater diversity compared to the RBM samples, consistent with $\pmfAnn$ better matching the data distribution.}
\label{fig:cd1-20-dbr}
\end{figure}

As discussed in Section~\ref{sec:raise-tractable}, RAISE is a stochastic lower bound on the log-likelihood of the annealing model $\pmfAnn$, 
but not necessarily of the RBM itself. 
When $\pmfAnn$ is a good approximation to the RBM, RAISE gives a conservative estimate of the RBM log-likelihood. 
However, it is possible for RAISE to overestimate the RBM log-likelihood if $\pmfAnn$ models the data distribution
better than the RBM itself, for instance if the approximation attenuates pathologies of the RBM.
We observed a single instance of this in our RBM experiments: the mnistCD1-20 RBM, 
with the data base rate initialization. As shown in Figure~\ref{fig:cd1-20-dbr}, 
the RAISE estimates exceeded the AIS estimates for small $\ndist$, and declined as $\ndist$ was increased. 
Since RAISE gives a stochastic lower bound on $\log \pmfAnn$ and AIS gives a stochastic upper bound on $\log \pmfTwo$,
this inversion implies that $\pmfAnn$ significantly outperformed the RBM itself. 
Indeed, the RBM (mistakenly) assigned 93\% of its probability mass 
to a single hidden configuration, while the RAISE model spreads its probability mass among more diverse configurations.

\begin{figure}[t]
\vspace{-0.1in}
\begin{minipage}{0.26 \textwidth}
\includegraphics[width=\textwidth]{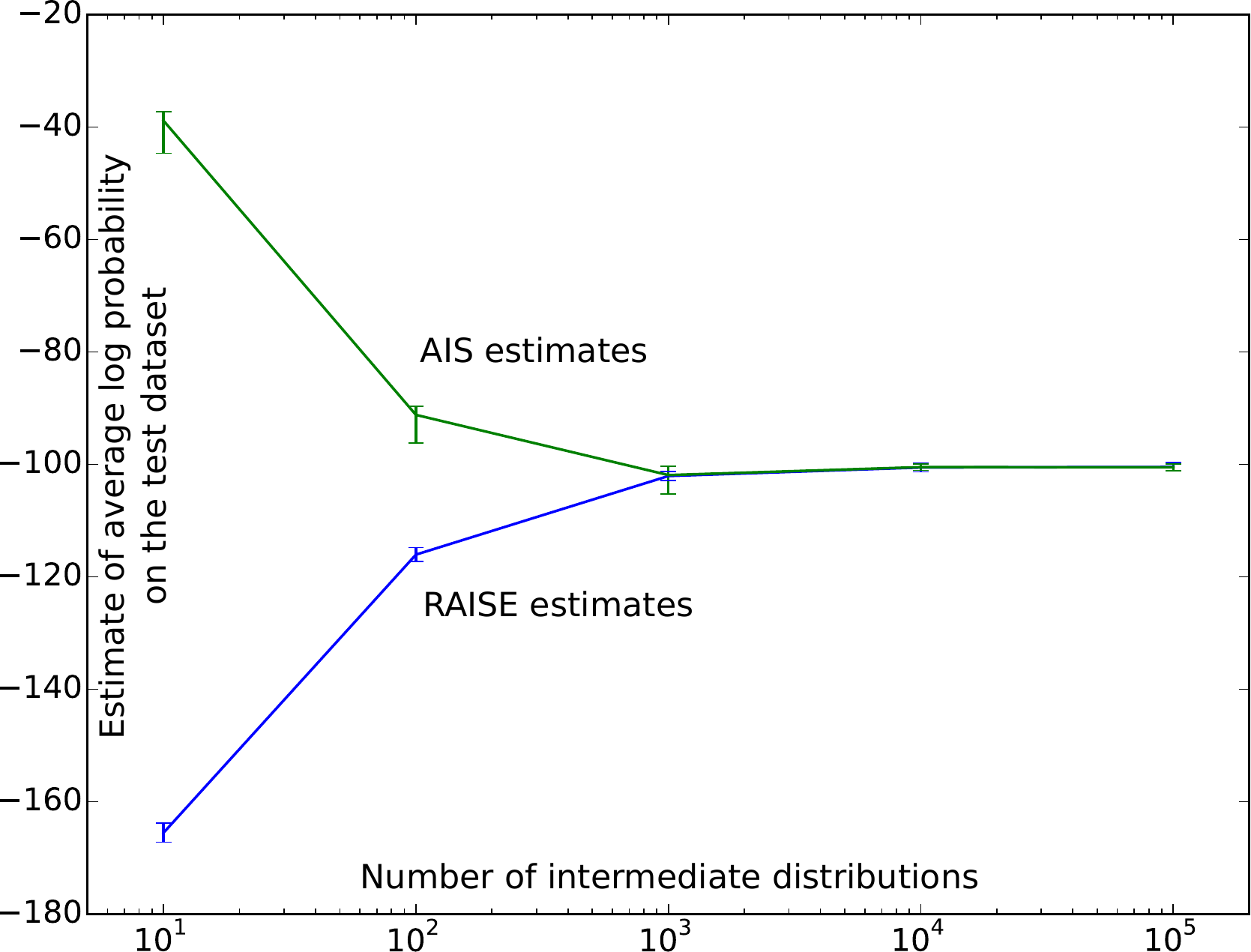}
\end{minipage}
\begin{minipage}{0.2 \textwidth}
\includegraphics[width=\textwidth]{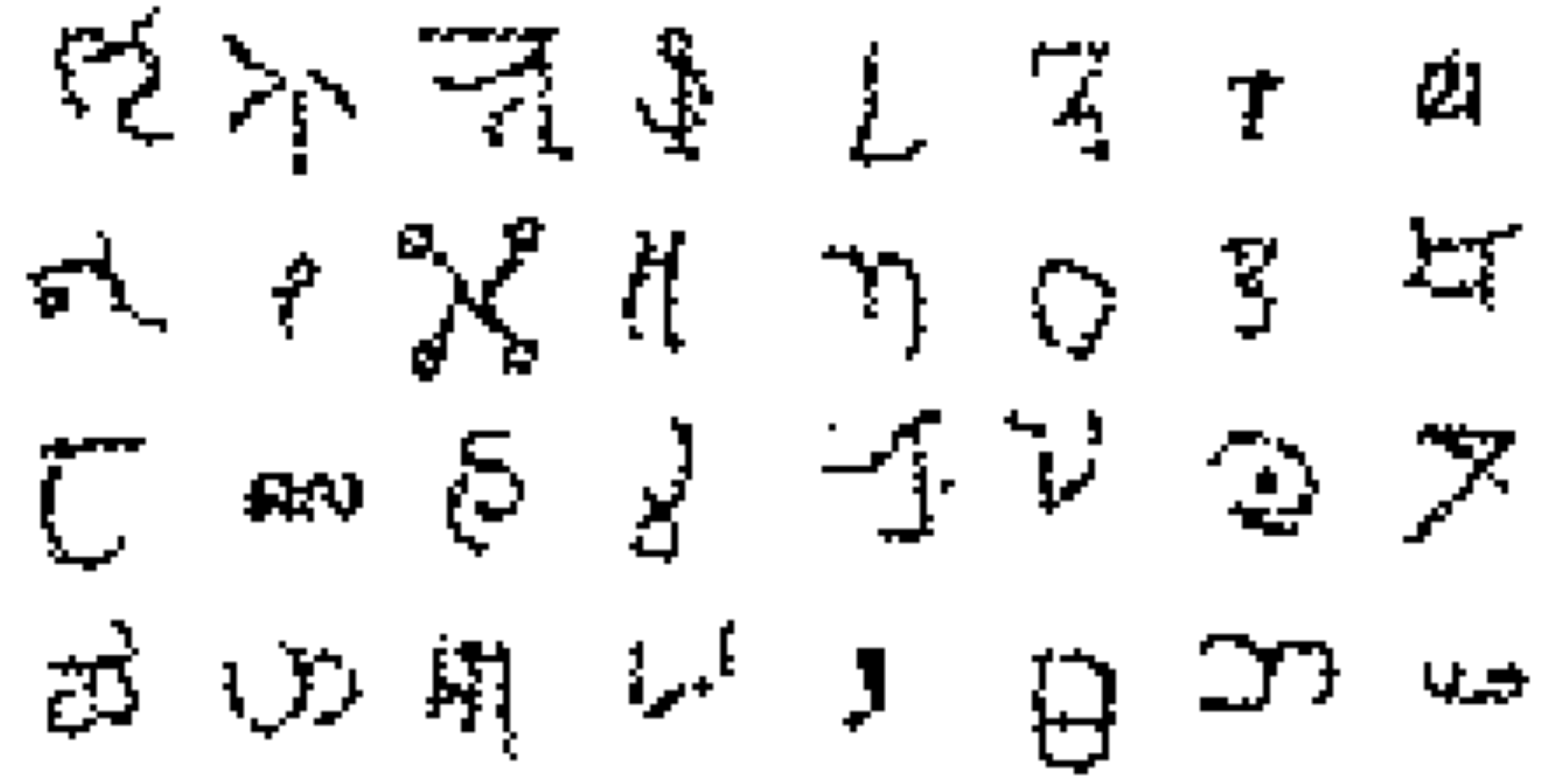}\\
\vspace{-0.5em}
\includegraphics[width=\textwidth]{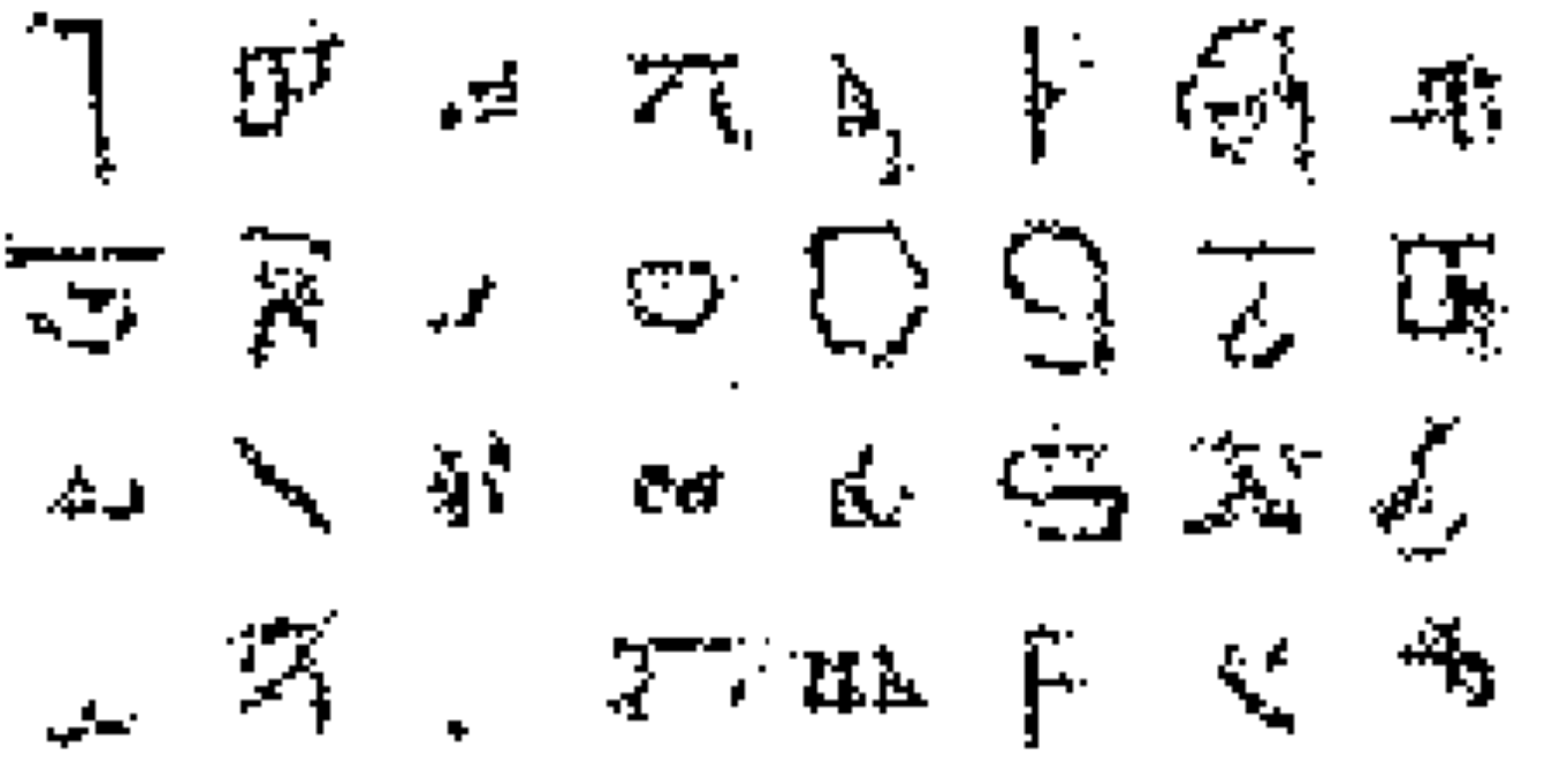}
\end{minipage}
\vspace{-0.1in}
\caption{\small {\bf Left: } AIS and RAISE estimates of omniPCD-1000 RBM average test log-probabilities
with annealing from a uniform initial distribution {\bf Top right: } 32 training samples from Omniglot training set {\bf Bottom right: } 32 independent samples from the omniPCD-1000 RAISE model with 100,000 
intermediate distributions.}
\label{fig:omni-pcd-1000}
\vspace{-0.1in}
\end{figure}

In all of our other RBM experiments, the AIS and RAISE estimates with DBR initialization and $\ndist=\textrm{100,000}$ agreed to within 0.1 nats.
Figure~\ref{fig:omni-pcd-1000} shows one such case, for an RBM trained on the challenging Omniglot dataset.

Overall, the RAISE and AIS estimates using DBR initialization agreed closely in all cases, and RAISE gave conservative estimates in all but one case, suggesting that RAISE typically gives accurate and conservative estimates of RBM test log-probabilities.

\subsection{Deep Boltzmann Machines}
\vspace{-0.05in}

We used RAISE to estimate the average test log-probabilities of two DBM models trained on MNIST and Omniglot.
The MNIST DBM has 2 hidden layers of size 500 and 1000, and the Omniglot DBM has 2 hidden layers each of size 1000. As with RBMs, 
we ran RAISE on 100 random test examples and used the DBM 
log unnormalized probability, $\log \pmfUnnorm(\visUnits)$, 
as a control variate.
To obtain estimates of the DBM unnormalized probability $\pmfUnnorm(\visUnits) = \sum_{\hidUnitsOne, \hidUnitsTwo} \pmfUnnorm(\visUnits, \hidUnitsOne, \hidUnitsTwo)$ we used simple importance sampling 
$\pmfUnnorm(\visUnits) = \expect_q{\left(\frac{\pmfUnnorm(\visUnits, \hidUnitsTwo)}{q(\hidUnitsTwo \given \visUnits)}\right)}$ 
with 500 samples, where the proposal distribution $q$ was 
the mean-field approximation to the conditional distribution $\pmf(\hidUnitsTwo \given \visUnits)$. 
The term $\pmfUnnorm(\visUnits, \hidUnitsTwo)$ was computed by 
summing out $\hidUnitsOne$ analytically, 
which is efficient because the conditional distribution 
factorizes.\footnote{Previous work \citep[\emph{e.g.}][]{dbm} estimated 
 $\log \pmfUnnorm(\visUnits)$ 
using the mean-field lower bound. We found importance sampling to give more accurate results in the context of AIS. 
However, it made less difference for RAISE, where the log unnormalized probabilities are merely used as a control variate.}

\begin{figure}
\includegraphics[width=0.23\textwidth]{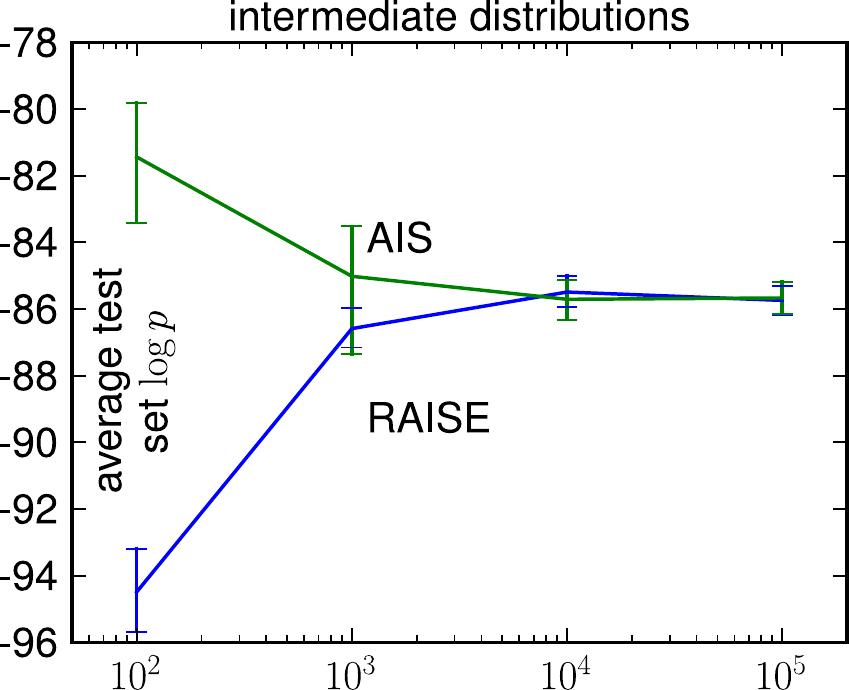}
\includegraphics[width=0.23\textwidth]{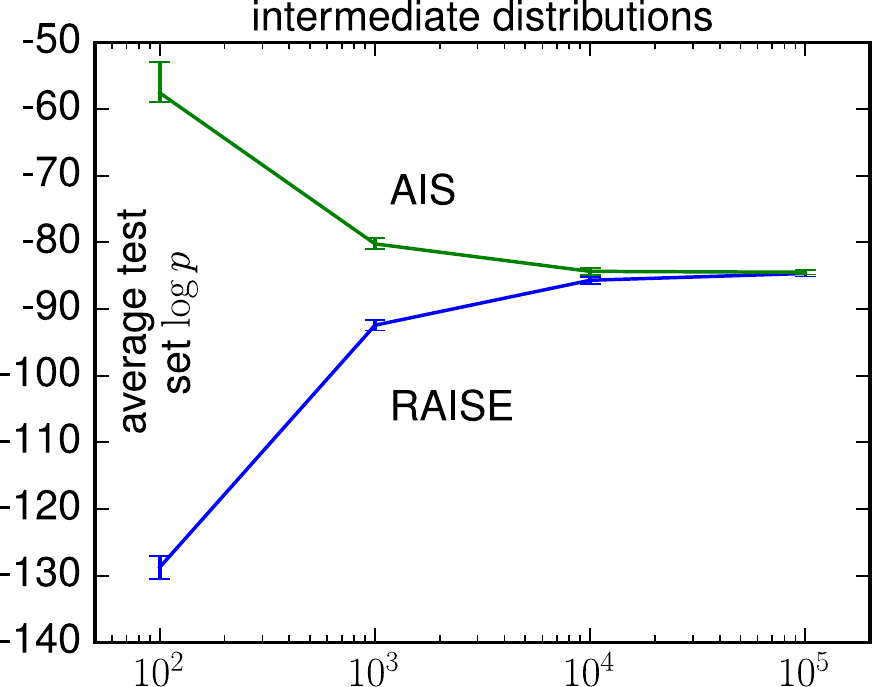}
\vspace{-0.1in}
\caption{\small Average test 
log-probability estimates for MNIST models as a function of $K$. {\bf Left:} the DBM.
 {\bf Right:} the DBN.}
\label{fig:dbn}
\vspace{-0.1in}
\end{figure}

We compared the RAISE estimates to those obtained using AIS. 
All results for $\ndist=\textrm{100,000}$ are shown in Table~\ref{tbl:deep-models-comparison}, and the estimates for the MNIST DBN are plotted as a function of $\ndist$ in Figure~\ref{fig:dbn}.
All estimates for the MNIST DBM with $\ndist=\textrm{100,000}$ agreed quite closely, which is a 
piece of evidence in favor of the accuracy of the estimates.
Furthermore, RAISE provided conservative estimates of log-probabilities for small $\ndist$, in contrast with AIS, which gave overly optimistic estimates.
For the Omniglot DBM, RAISE overestimated the DBM log-probabilities by at least 6 nats, implying that the annealing model fit the data distribution better than the DBM, analogously to the case of the mnistCD1-20 RBM discussed in Section~\ref{sec:rbm-experiments}. 
This shows that RAISE does not completely eliminate the possibility of overestimating an MRF's test log-probabilities.

\begin{table}[t]
\vspace{-0.1in}
\begin{scriptsize}
\begin{tabular}{l|rrr|rrr}
 \multicolumn{1}{c}{} & \multicolumn{3}{c}{uniform}            & \multicolumn{3}{c}{data base rates} \\
Model           & RAISE    & AIS      & gap      & RAISE     & AIS      & gap        \\ \hline
MNIST DBM       & -85.69   & -85.72   & -0.03    &  -85.74   & -85.67   & 0.07     \\
Omniglot DBM    & -104.48  & -110.86  & -6.38    & -102.64   & -103.27  & -0.63           \\
MNIST DBN       & -84.67   & -84.49   & 0.18     &  ---      & ---      & ---   \\
Omniglot DBN    & -100.78  & -100.45  & 0.33     &  ---      & ---      & ---
\end{tabular}
\end{scriptsize}
\vspace{-0.1in}
\caption{\small Test log-probability estimates for deep models with $\ndist=\textrm{100,000}$. {\bf gap:} the difference $\textrm{AIS} - \textrm{RAISE}$}
\label{tbl:deep-models-comparison}
\end{table}

\subsection{Deep Belief Networks}
\vspace{-0.05in}

In our final set of experiments, 
we used RAISE to estimate the average test log-probabilities of DBNs trained on MNIST and Omniglot. The MNIST DBN had two hidden layers of size 500 and 2000, and the Omniglot DBN had two hidden layers each of size 1000.
For the initial distribution $\pmf_0$ we used the uniform distribution, as the DBR distribution is not
defined for DBNs.
To obtain estimates of DBN unnormalized probabilities 
$\pmfUnnorm(\visUnits)=\sum_{\hidUnitsOne}{\pmf(\visUnits \given \hidUnitsOne)\pmfUnnorm(\hidUnitsOne)}$ 
we used importance sampling 
$\pmfUnnorm(\visUnits)=\expect_q\left(\frac{\pmf(\visUnits \given \hidUnitsOne)\pmfUnnorm(\hidUnitsOne)}
{q(\hidUnitsOne \given \visUnits)}\right)$ with 500 samples, where $q$ was the DBN recognition distribution \citep{dbn}. 

All results for $\ndist=\textrm{100,000}$ are shown in Table~\ref{tbl:deep-models-comparison}, and 
Figure~\ref{fig:dbn} shows the estimates for the MNIST DBN as a function of $\ndist$. For both DBNs,
RAISE and AIS agreed to within 1 nat for $\ndist = \textrm{100,000}$, and RAISE gave conservative
log-probability estimates for all values of $\ndist$.

\subsection{Summary}
\vspace{-0.05in}
Between our RBM, DBM, and DBN experiments, we compared 10 different models using both uniform and 
data base rate initial
distributions. In all but two cases (the mnistCD1-20 RBM and the Omniglot DBN), RAISE gave estimates at or below
the smallest log-probability estimates produced by AIS, suggesting that RAISE typically gives conservative estimates.
Furthermore, in all but one case (the Omniglot DBM), the final RAISE estimate agreed with the lowest AIS estimate
to within 1 nat, suggesting that it is typically accurate.

\section{Conclusion}
\vspace{-0.1in}
In this paper we 
presented the Reverse AIS Estimator (RAISE), 
which gives a stochastic lower bound on the log-likelihood of an 
approximation to an MRF model.
Our experimental results show that  
RAISE typically produces accurate, yet conservative, estimates of log-probabilities for RBMs, DBMs, and DBNs.
More importantly, by using RAISE and AIS in conjunction, one can judge the accuracy of one's results by measuring the agreement of the two estiatmators. RAISE is simple to implement, requiring only the same transition operators as AIS,
so it gives a simple and practical way to evaluate MRF test log-probabilities.

\section*{Acknowledgements}
\vspace*{-10pt}
This research was supported by NSERC, Google, and Samsung.

\bibliographystyle{plainnat}
\bibliography{reverse_annealing}

\end{document}